# A Convolutional Autoencoder for Multi-Subject fMRI Data Aggregation

**Po-Hsuan Chen**[1], **Xia Zhu**[2], **Hejia Zhang**[1], **Javier S. Turek**[2], **Janice Chen**[3],
**Theodore L. Willke**[2], **Uri Hasson**[3], **Peter J. Ramadge**[1]
[1]Department of Electrical Engineering, Princeton University, [2]Intel Labs,
[3]Princeton Neuroscience Institute and Department of Psychology, Princeton University

## Abstract

Finding the most effective way to aggregate multi-subject fMRI data is a long-standing and challenging problem. It is of increasing interest in contemporary fMRI studies of human cognition due to the scarcity of data per subject and the variability of brain anatomy and functional response across subjects. Recent work on latent factor models shows promising results in this task but this approach does not preserve spatial locality in the brain. We examine two ways to combine the ideas of a factor model and a searchlight based analysis to aggregate multi-subject fMRI data while preserving spatial locality. We first do this directly by combining a recent factor method known as a shared response model with searchlight analysis. Then we design a multi-view convolutional autoencoder for the same task. Both approaches preserve spatial locality and have competitive or better performance compared with standard searchlight analysis and the shared response model applied across the whole brain. We also report a system design to handle the computational challenge of training the convolutional autoencoder.

## 1 Introduction

There is growing interest in finding more effective ways of aggregating multi-subject fMRI data. This is an important problem both for reasons of scientific generality and for attaining higher statistical sensitivity in complex fMRI studies. The standard method for aggregating such data uses anatomical registration across subjects [1, 2, 3]. Since this does not adequately align subjects' functional responses [3, 4, 5, 6], it is usually followed by spatial smoothing to blur differences in subject functional responses. There is a growing body of recent research exploring more direct approaches to functional registration. This includes cortical warping to align functional time series [4] and functional connectivity across subjects [5, 7], and the application of factor methods such as ICA, CCA, IVA, hyperalignment (HA) [8, 9, 10, 11, 12, 13, 6]. The most recent work in this vein, called the shared response model (SRM) [14], has focused on learning probabilistic latent factors that jointly model subject specific functional topographies and a shared temporal response.

Factor models operate on the principle of aggregating information across one or more dimensions of the data (space, time, subject). For example, HA and SRM aggregate information across space and subjects. Since aggregating across space (voxels) reduces anatomical spatial locality, these methods are usually applied to large pre-selected regions of interest (ROI), e.g., ventral temporal cortex [6] and posterior medial cortex [14]. Applying the models in this way can yield significant gains over prior methods in identifying informative responses in pre-selected regions [14]. However, these methods suffer from an important limitation: a lack of spatial locality. That is, all voxels within the selected region may contribute to the measure that is ultimately derived (e.g., a classification score). This limitation is at odds with a fundamental goal of neuroscience, which is to determine how local brain regions are associated with specific cognitive functions. For example, ventral temporal cortex is known to contain a multitude of sub-areas, each with its own specialized function [15]. If all of these sub-areas enter an analysis together, overall classification scores may improve, but the ability to make inferences about the functional properties of individual sub-areas is lost.

Here we focus on the preservation of spatial locality during whole brain multi-subject data aggregation, with the aim of improving anatomical and functional interpretability of the analysis results. By preserving spatial locality we mean that information is only aggregated in a small region (e.g. ball) about each voxel. A natural approach is that can satisfy this constraint to combine factor models and searchlight based analysis [16, 17]. Searchlight analysis uses a small window of contiguous voxels around a known location to conduct a spatially local analysis. This analysis is performed at all locations in the volume, thus generalizing an ROI approach to multiple (overlapping) spatially local "searchlights" across the brain. To handle $m$ subjects, the analysis can be performed across $m$ linked and co-centered searchlights, one per subject. This provides multi-subject, local data aggregation tailored to each searchlight [18]. In this paper we focus on this approach with an aim of making a connection between searchlight analysis and convolution neural networks. Other approaches that aim to ensure spatial locality are also possible. For example, a data-driven approach that learns "soft" boundaries of local activated areas.

We explore the application of searchlights in two distinct ways: by combining the SRM with searchlights (S-SRM) and by using a multi-view convolutional autoencoder (CAE). A search light version of SRM is not conceptually new. We bring it in as a fairer benchmark for CAE than factor models without spatial locality constraints. To understand the relevance of a convolutional autoencoder we first note that a two layer fully connected autoencoder can replicate the performance of SRM on multi-subject fMRI data. But like the SRM, this autoencoder does not have spatial locality. We then argue that we can add spatial locality by transitioning from a fully connected to a convolutional autoencoder. To see this consider the post-training application of a S-SRM analysis and that of a single layer convolutional neural network (CNN). In a S-SRM analysis, a fixed sized window is moved over the data and at each location we form $k$ inner products between weight vectors (learned functional topographies in [14]) and the windowed data. Similarly, in a convolution using $k$ filters, a fixed size filter support is moved over the data and at each location $k$ inner products of filter coefficient vectors and the windowed data are computed. In both cases the results are recorded and indexed by the coordinates of the region center. Subsequent analysis is then based on the outputs produced in each case. While the above analogy shows a clear similarity, the two approaches also differ in important ways. First, in a S-SRM the weight vectors can depend on the searchlight index but in a CNN the filter weights are invariant with location. Thus the searchlight approach has a key advantage: it can vary data aggregation depending on anatomical location. Second, the SRM (and many other factor methods) impose a nonlinear geometric constraint on the weight vectors (e.g., orthonormality), whereas CNN filter weights are not directly geometrically unconstrained except perhaps in norm. Third, a CNN contains distributed nonlinear activation functions whereas in a factor model, data factorization is a global nonlinear operation. It is well known, however, that a fully connected neural network can make use of its distributed activation functions to approximate nonlinear functions [19, 20].

There are several previous applications of deep learning to fMRI data. We review this recent literature and draw connections wth our proposal. For unsupervised feature extraction the $l_1$ regularized restricted Boltzmann machine has demonstrated comparable performance with ICA while giving more localized features [21]. 1D temporal convolutional autoencoders have been applied on fMRI data in matrix form (voxel-by-time) in a temporal convolutional neural network framework [22]. A recent work on classifying neuroimaging data used semi-supervised linear autoencoders to learn compressed representations of the neuroimaging data in the unsupervised stage [23]. Another classification paper uses deep neural networks to perform supervised learning, with the fMRI data as input and the corresponding class labels as output [24]. In this work, the intensities of voxels in the each anatomical region of interest are averaged to help deal with the variability across subjects. These previous applications of deep learning to fMRI data do not explore the co-activation across subjects nor the preservation of spatial locality in the aggregation of multi-subject data.

Our goal then is to design a multi-layer convolutional autoencoder for multi-subject, whole brain, spatially local, fMRI data aggregation. To do so we create a network structure that matches the inherent multi-dataset nature of the problem and address some computational challenges arising from dealing with large-scale, multi-subject fMRI data. Our key contribution is to show that a suitably designed convolutional autoencoder can provide data aggregation that is competitive with methods based on whole brain searchlight analysis using latent factor methods. We also examine approaches to address the computational challenges of training a convolutional autoencoder using multi-subject fMRI data.

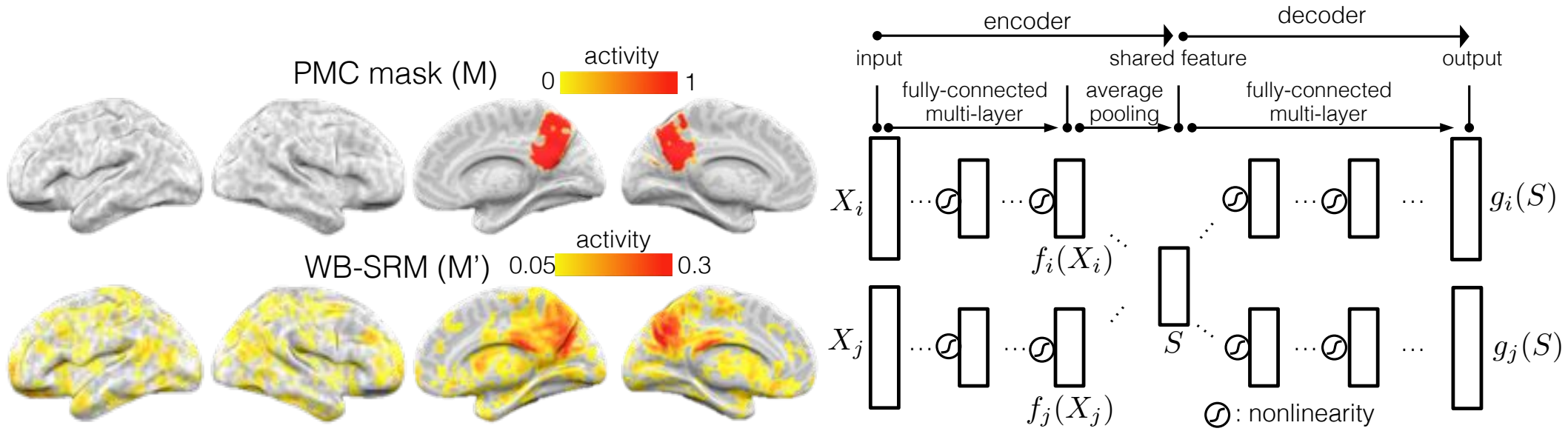

Figure 1: Left: Illustration of the lack of spatial locality of whole brain SRM (WB-SRM) analysis. Right: A nonlinear, fully connected autoencoder that can match the performance of SRM but also lacks spatial locality.

## 2 Limitations of Current Methods

fMRI time-series data $X_i \in \mathbb{R}^{v_x \times v_y \times v_z \times d}$, $i = 1{:}m$, is collected from $m$ subjects presented with an identical, temporally synchronized stimulus. The dimensions $(x, y, z)$ of $X_i$ are spatial coordinates in the brain, $(v_x, v_y, v_z)$ are the number of voxels in the $(x, y, z)$ dimensions, and $d$ is the number of time samples in units of repetition time (TR). $X_i$ can be regarded as a 4D tensor, but to afford wider accessibility we use standard multivariate notation. Our objective is to model, across the whole brain, the elicited response shared by the subjects while preserving its spatial locality. To do so we set out to identify local subject specific patterns that co-activate in time across subjects.

The SRM approach proposed in [14] aims to achieve this goal by learning subject specific matrices $W_i \in \mathbb{R}^{v \times k}$, each with $k$ orthonormal columns, and a shared response $S \in \mathbb{R}^{k \times d}$ to minimize the reconstruction error $\sum_{i=1}^{m} \frac{1}{m} \|X_i - W_i S\|_F^2$. Once learning is complete, one can project held out data $X_i'$ for subject $i$ into the shared response space by computing $k$ inner products $S_i' = W_i^T X_i'$. One can also project this data into the voxel space of subject $j$ by computing $W_j W_i^T X_i'$. The imposed orthonormality constraint plays a key role in achieving the performance reported in [14]. If this constraint is removed, performance drops (see Sup. Mat.). In addition, if spatial locality is desired, it must be externally imposed by restricting the SRM domain to a spatially local ROI. Applying the method across the whole brain forgoes spatial locality. We can demonstrate this using the *sherlock* dataset (see §4). After using the dataset to learn the SRM on the whole brain we obtain $W_i \in \mathbb{R}^{v \times k}$, $i = 1{:}m$ and a shared response $S \in \mathbb{R}^{k \times d}$. We then create a synthetic brain map $M$ in the voxel space of subject 1 taking value 1 in a post medial cortex (PMC) anatomical ROI and 0 elsewhere, and use the learned matrices $W_1, W_2$ to map $M$ into the voxel space of subject 2: $M' = W_2 W_1^T M$. Preserving spatial locality requires that the support of $M'$ is close to that of $M$. The result (Fig. 1) clearly shows that special locality is not preserved.

Our problem can also be conceived as multi-view learning problem and in this context, fully connected neural networks and autoencoders have proven useful [25, 26, 27]. It is possible to connect the SRM and a linear autoencoder by simply removing the constraint $W_i^T W_i = I_k$ and viewing the SRM objective as the reconstruction loss of a fully-connected linear, single hidden layer autoencoder (see Fig. 1, Sup. Mat.). But dropping the above constraint reduces performance. In contrast, a nonlinear, multi-view autoencoder with two hidden layers (Fig. 1) can match the performance of SRM (details in Sup. Mat.). However, like SRM, this autoencoder does not preserve spatial locality. Nevertheless, it suggests a novel approach to the fMRI data aggregation problem.

## 3 A Convolutional Autoencoder for Multi-subject Data Aggregation

Motivated by the desire for spatial locality, we propose and investigate a 4D convolutional autoencoder (Fig .2) for multi-subject fMRI data aggregation. For simplicity, Fig. 2 shows only two subjects and we explain its operation in this context. The input data consists of 4D tensors $X_i, i = 1{:}m$ and the first layer is a 3D convolutional layer. To account for functional variability between subjects, this layer learns $k_1$ subject specific filters. However, filters with the same index are linked across the subjects. The output of the first layer is a set of $mk_1$ 3D feature maps $X_i^j$; one per subject $i$ and filter index $j$. As shown in Fig. 2, these are subject-grouped for each linked filter index. The grouped feature maps specify the activity level across subjects of linked local filters at locations across the brain. The second layer is average pooling across subjects. This identifies local patterns that co-activate across subjects (co-activating patterns). The result is $k_1$ shared feature maps. We do not expect activation of local spatial patterns alone to be informative of a shared response. Hence a second round of $k_3$ convolutions is performed over the $k_1$ activation patterns to identify local combinations of spatial activity patterns. This also introduces a second non-linearity into the network which is known to increase representational power [28]. The second convolutional layer computes $k_3$ 1D convolutions resulting in $k_3$ 3D feature maps. This design satisfies our goal of preserving spatial locality by aggregating the information across subjects from voxels within the filter support size. Each location in the final shared feature maps (L3) corresponds to brain searchlights linked across subjects. Finally, we use a single layer of convolution to generate the reconstructed datasets $\hat{X}_i = h_{i,\theta}(X_1, \ldots, X_m)$, where $\theta$ is the model parameters. These represent the manifestation of the shared response in each subject's brain.

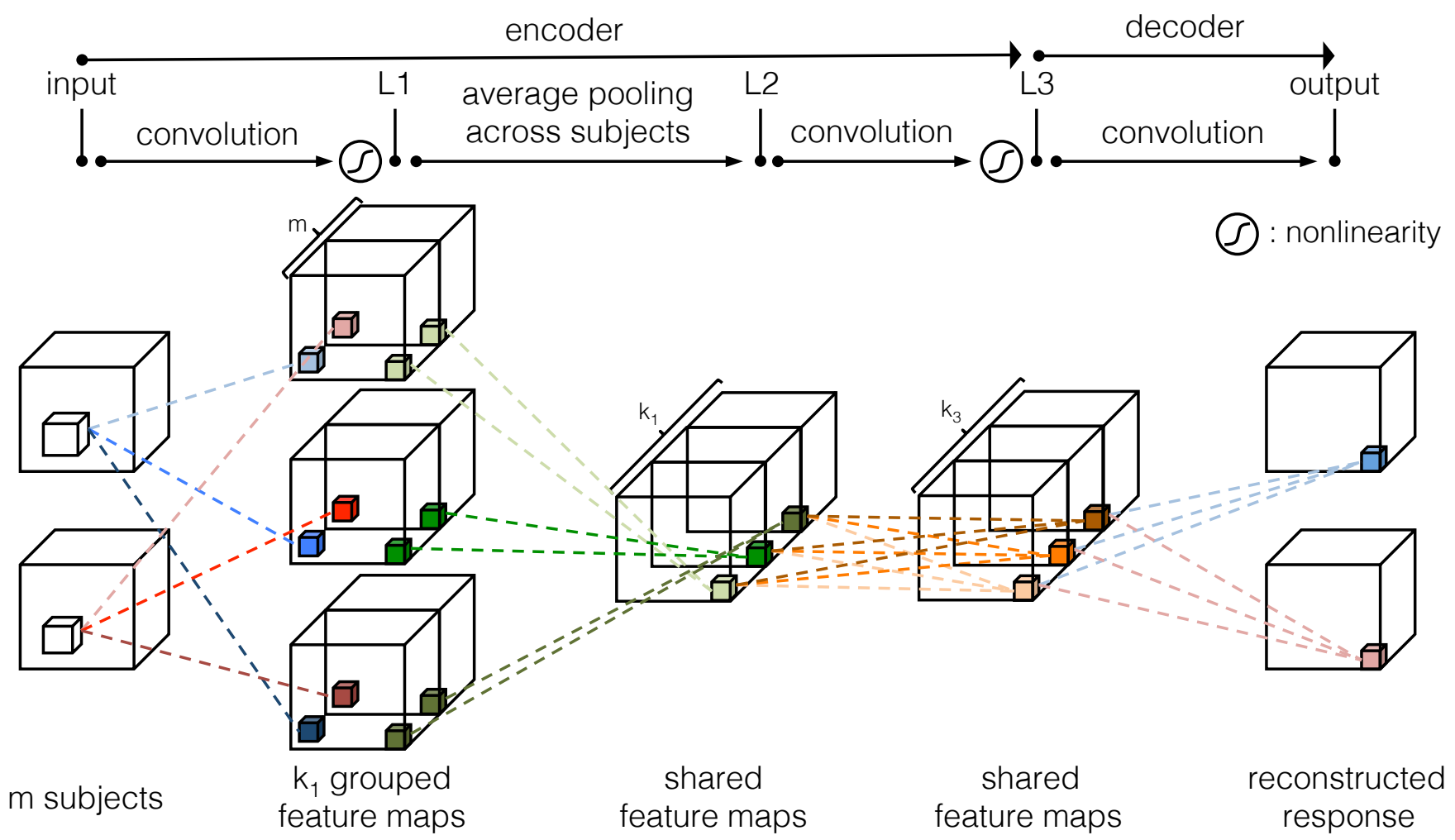

Figure 2: Proposed 4D Convolutional Autoencoder.

We train the convolutional autoencoder by minimizing the loss function

$$L(\theta; X) = \frac{1}{m}\sum_{i=1}^{m} \|X_i - h_{i,\theta}(X_1, \ldots, X_m)\|_F^2 + \lambda D_{\mathrm{KL}}(\rho\|\hat{\rho}). \quad (1)$$

The first term is the mean squared error between the reconstructed output $h_{i,\theta}(X_1, \ldots, X_m)$ and the subject's data; the second term is the Kullback-Leibler (KL) divergence to a binomial distribution with parameter $\rho$ [29]: $D_{\mathrm{KL}}(\rho\|\hat{\rho}) = \rho\log(\frac{\rho}{\hat{\rho}}) + (1-\rho)\log(\frac{1-\rho}{1-\hat{\rho}})$ with $\rho$ the desired sparsity and $\hat{\rho}$ the mean sparsity of the activation in the layer. This regularizes the network by sparsifying the $k_3$ shared feature maps in layer 3. We use the hyperbolic tangent activation function since the data is z-scored and it yields shared feature maps with positive and negative values. The sparsity regularization is computed by scaling and shifting the hyperbolic tangent output to $[0, 1]$. Dropout is used to reduce overfitting [30]. We select the parameters $\rho$ and $\lambda$ using cross-validation and fix dropout on hidden layers to the typical value of $0.5$ [30] and deactivate it on the input layer.

In our convolutional autoencoder, the number of model parameters is much smaller than the number of activations. Therefore, we adopt a data parallel method for distributed training that reduces communication overhead. We implement a distributed training framework for Theano [31] based on a synchronous Stochastic Gradient Descend (SGD) [32, 33] to handle the computational load for training the network. We select a synchronous method over asynchronous SGD because of the better convergence properties [33]. The synchronous SGD method has many processes running in parallel, each maintaining a copy of the entire model. Every SGD iteration, a mini batch is assigned to each process to compute a local gradient. Then, all these gradients are aggregated by a binomial reduction tree based collective operation. Eventually, the local models are updated using the aggregated gradient. In addition, we initialize all filters in the first layer with values from a random orthogonal matrix. We use RMSprop [34] to adaptively adjust the learning rate. For decay rate and smoothing value, we swept in the range {0.9, 0.99, 0.999} and $\{10^{-4}, 10^{-6}, 10^{-8}\}$, respectively. The initial learning rate depends on the batch size and the number of nodes used.

| Dataset | TRs (s/TR) | Voxel Region | # Voxels |
|---|---|---|---|
| audiobook (narrated story) [35] | 449(2) | whole brain (WB) in MNI [3] | 70273 |
| sherlock-movie (audio-visual movie) [36] | 1976(2) | whole brain (WB) in MNI [3] | 70273 |
| | | posterior medial cortex (PMC)[37] | 813 |
| sherlock-recall (movie free-recall) [36] | 437∼1009(2) | whole brain (WB) in MNI [3] | 70273 |
| | | posterior medial cortex (PMC)[37] | 813 |

Table 1: fMRI datasets are shown in the left two columns, and the ROIs are shown in right two columns. We use 9 subjects from version of datasets that match the data in the corresponding publications.

# 4 Experiments and Results

The performance of S-SRM and the CAE was evaluated using two fMRI datasets (Table 1). These were collected using a 3T Siemens Skyra scanner using different subjects and preprocessing pipelines. The *sherlock* dataset contains an audio-visual portion, *sherlock-movie*, collected while subjects watched a 50 min. section of an episode of the BBC series "Sherlock", and a recall portion, *sherlock-recall*, in which subjects verbally recalled out loud the episode in as much detail as possible (without any experimenter guidance or cues). The *audiobook* dataset was collected while subjects listened to a 15 min. narrated story. The primary metrics are prediction accuracy, used as a proxy for relevant information, and spatial locality. The high accuracy regions indicate the strongest presence of information relevant to the testing hypothesis.

For the CAE we use a $5 \times 5 \times 5$ support region in first layer convolutions for full resolution fMRI data and $3 \times 3 \times 3$ regions for data down-sampled by 2. After training, held out data is mapped from the input layer to the shared response (bypassing across subject pooling) at the output of layer 3. S-SRM uses searchlights sized as above. Each searchlight contains $v_s$ voxels and we use $k = 10$ features per searchlight. The training data for voxels in the $m$ across subject linked searchlights are used to learn a SRM and the learned subject-specific maps $W_i \in \mathbb{R}^{v_s \times k}$ are used to project held out data into the shared space for each searchlight. For both the CAE and S-SRM we then conduct time segment matching and brain map matching experiments using the data of a held out subject. The resulting accuracy of the matching tasks is assigned to the center voxel of the corresponding input region for CAE or the searchlight for S-SRM. This enables us to plot a local accuracy map across the whole brain. In comparisons with standard searchlight based analysis we use the same sized searchlights and for comparisons with single region SRM (whole brain or ROI) we use $k = 100$ features (for details see Sup.Mat.).

For training the CAE, we use the distributed synchronous SGD described in §3, applying MPI parallelism at the mini batch level. In addition, we tune Theano to make full use of OpenMP. Moreover, certain operations in NumPy and SciPy versions run serially and therefore we develop NumPy extension modules in C++ parallelized with OpenMP to speed them up. With these optimizations, we obtain up to 67× training speedup on a single node comparing to the original Theano version, and another 7× speedup on an 8-node CPU cluster. Furthermore, we only load necessary data according to the mini batch to maintain a reduced memory footprint in each process.

We run our experiments on an 8-node cluster[1], interconnected by an Arista 10GE switch. Each node of the cluster has a motherboard with 2 Intel ® Xeon ® E5-2670 processors [2], both running at 2.6GHz, and with 256GB memory. The convolutional autoencoder is implemented in Python, with OpenMP for multi-threading, and mpi4py for multi-node parallelism. The software packages used in our experiments include Intel optimized Theano [31] (version rel-0.8.0rc1). We use the Anaconda distribution of Python with the following packages: Intel ® MKL 11.3.1, NumPy 1.10.4 and SciPy 0.17.0.

**Exp. 1: Local region time segment matching.** We first use the *sherlock-movie* and *audiobook* datasets to replicate an experiment from [14]. The experiment compares a standard searchlight analysis (SL) with S-SRM and the CAE. For each dataset, the movie data was split into halves, one half was used for training the other for testing; then the roles were reversed and results averaged. The experiment tests if a 9 TR time segment from the testing data of a held-out subject can be located in the testing data of the subjects used in training. In the testing phase, we map subject's testing data from the input to the shared feature map (without conducting average pooling across subjects). A random 9 TR test segment from the testing half of the held out subject's data is projected onto the shared space and we locate this segment in the averaged shared response of the other subject's testing data by maximizing Pearson correlation. Segments overlapping with the test segment are excluded from the matching process. We record average accuracy and standard error by two-fold cross-validation over the data halves and leave-one-out over subjects. Each dataset is in MNI space [3]. The accuracy maps are shown on the left of Fig. 3. Accuracies below 0.05 were set to zero. Since each searchlight contains only a small local view, its predictive performance is expected to be low. The experiment was also conducted using a univariate voxel test but no voxel scored above 0.05. Chance accuracy is 0.0044 for the *audiobook* dataset and 0.001 for *sherlock-movie* dataset.

[1] Software and workloads used in performance tests may have been optimized for performance only on Intel microprocessors. Performance tests, such as SYSmark and MobileMark, are measured using specific computer systems, components, software, operations and functions. Any change to any of those factors may cause results to vary. You should consult other information and performance tests to assist you in fully evaluating your contemplated purchases, including the performance of that product when combined with other products. For more information go to http://www.intel.com/performance

[2] Intel and Xeon are trademarks of Intel corporation in the U.S. and/or other countries.

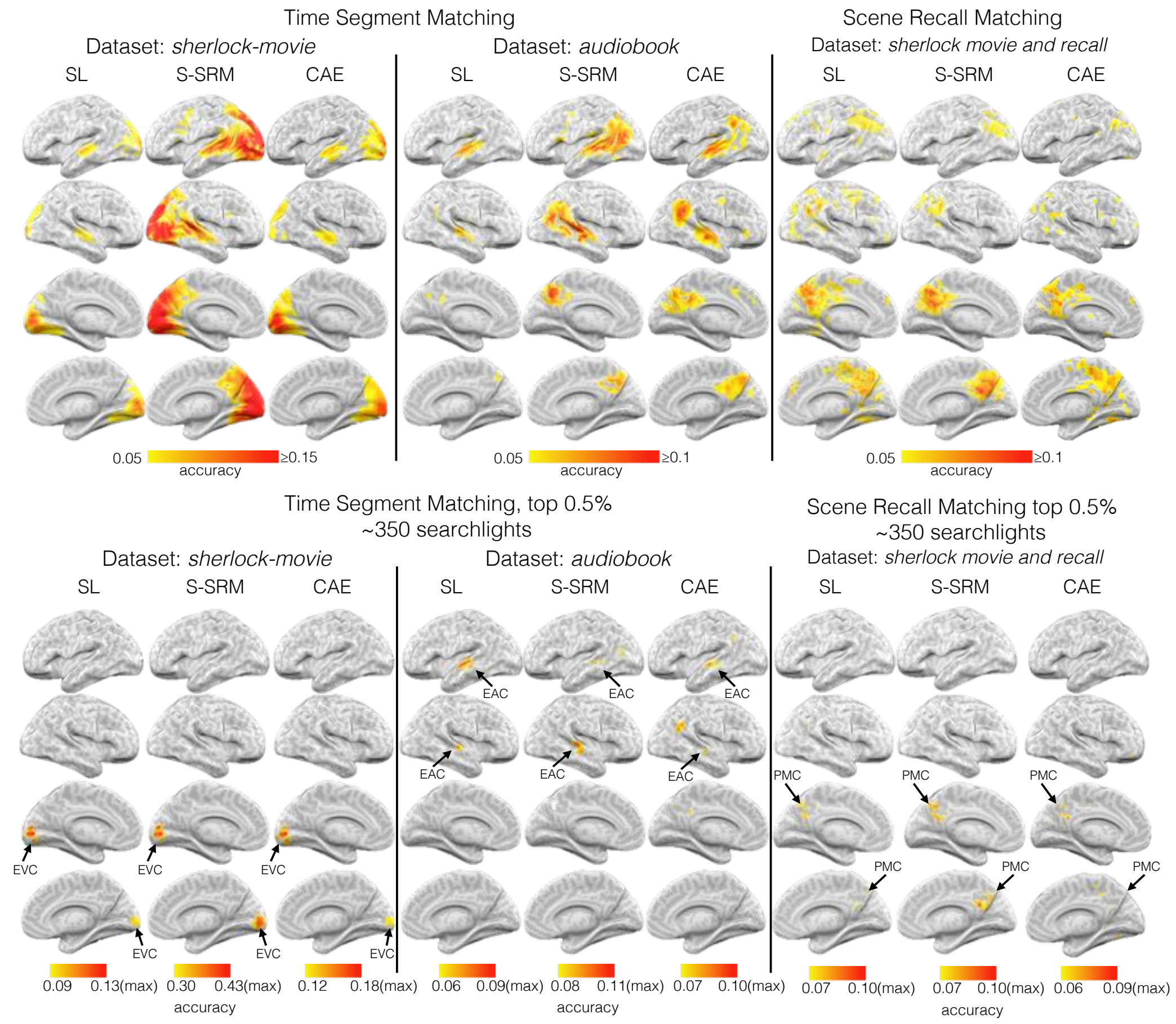

Figure 3: Top Left: Accuracy maps for Exp. 1 using *sherlock-movie* and *audiobook*; Top Right: Accuracy maps for Exp. 2 using *sherlock-movie* and *sherlock-recall*. Top figures are thresholded at corresponding scales for visualization clarity purpose. Please refer to bottom row figures for high end of the range. Bottom Left: Accuracy maps for top 0.5% searchlights for Exp. 1; Bottom Right: Accuracy maps of top 0.5% searchlights for Exp. 2. Early Visual Cortex (EVC), Early Auditory Cortex (EAC), Posterior Medial Cortex (PMC).

**Exp. 2: Scene recall matching.** We now use *sherlock-movie* and *sherlock-recall* to compare standard SL analysis, S-SRM and CAE analyses on a more challenging task. We label each TR of the *sherlock-recall* data with the corresponding scene based on the subject's verbal description. The TRs captured during a subject's recall of the same scene are averaged. Our goal is to test if subjects have a similar brain activation pattern when retrieving memory of the same scene. To do so we attempt to classify the scene of the recall responses of a left out subject . The whole movie is used to train S-SRM and the CAE. The effectiveness of the learnt shared response is then tested using data from a held out subject. After projecting the *sherlock-recall* data to the shared space, an SVM classifier is trained and the average classifier accuracy and standard error is recorded by leave-one-out across subject testing. The results are shown as the accuracy plots on the right in Fig. 3. Chance accuracy is $0.02$.

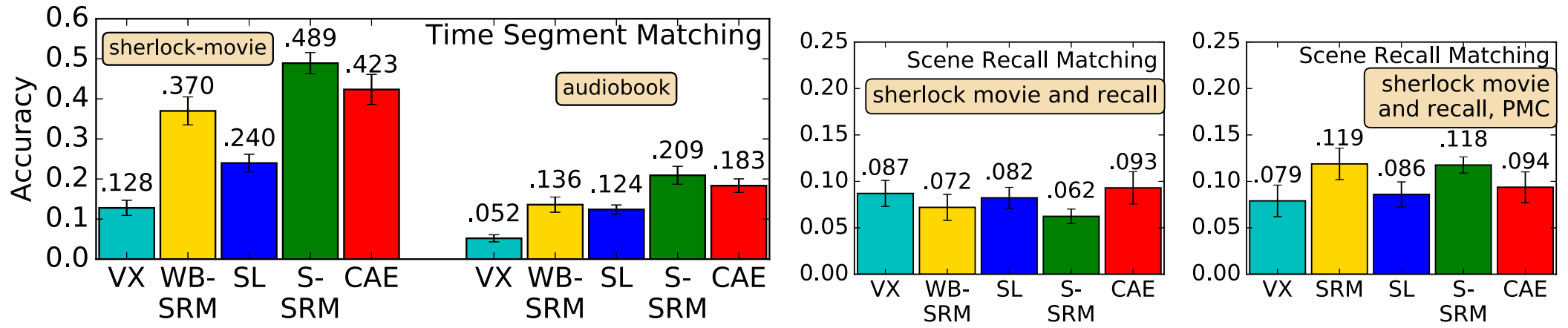

Figure 4: Prediction accuracies for Exp. 3. Left: Comparison of 9 TRs time segment matching on two datasets. Middle: Comparison of movie scene recall classification on *sherlock*. Right: Comparison of movie scene recall classification on *sherlock* in PMC. Error bars: $\pm 1$ stand. error.

**Exp. 3: Whole brain time segment and scene recall matching.** In this experiment we investigate how well we can perform time segment matching and scene recall matching using a classifier that combines locally learnt information across the whole brain. This experiment compares five approaches: whole brain voxel analysis (VX), whole brain SRM (WB-SRM) with $k = 100$ features, standard SL, S-SRM, and CAE. The experiment procedure is similar to Exp. 1 and Exp. 2, however, instead of doing classification in each local region, we classify using the results of all the local analyses across the whole brain. Whole brain voxel analysis (VX) is done by directly calculating time segment matching on whole brain voxel data without any model. WB-SRM is done by applying SRM ($k = 100$) on whole brain data. Standard SL, S-SRM and CAE are local methods applied as before directly to whole brain data. By aggregating the local information from all local regions, we expect higher predictive power. The results are shown in the left plot of Fig. 4 for time segment matching and in the middle plot for scene recall. Motivated by these results for scene recall we also conducted the same scene recall experiment in the PMC ROI. These results are shown in the right plot of the figure.

**Exp 4: Dispersion** We now examine how well S-SRM and the CAE address the issue of spatial locality. We conduct the same experiment described in §2 and shown in Fig. 1. We use two ROI regions to compare the spatial dispersion of S-SRM and CAE (Fig. 5). As expected, S-SRM and the CAE have much less spatial dispersion than WB-SRM. The S-SRM exhibits slightly greater dispersion than the CAE.

## 5 Discussion and Conclusion

Our objective is to accomplish whole brain, multi-subject, fMRI data aggregation while preserving the spatial locality of information. The dispersion experiment (Fig. 5) indicates that both S-SRM and the CAE preserve spatial locality. The key remaining issue is whether aggregation of fMRI responses using these methods better distinguishes local and global cognitive states. To check this we use the accuracy maps as a proxy measure of the effectiveness of the information aggregation.

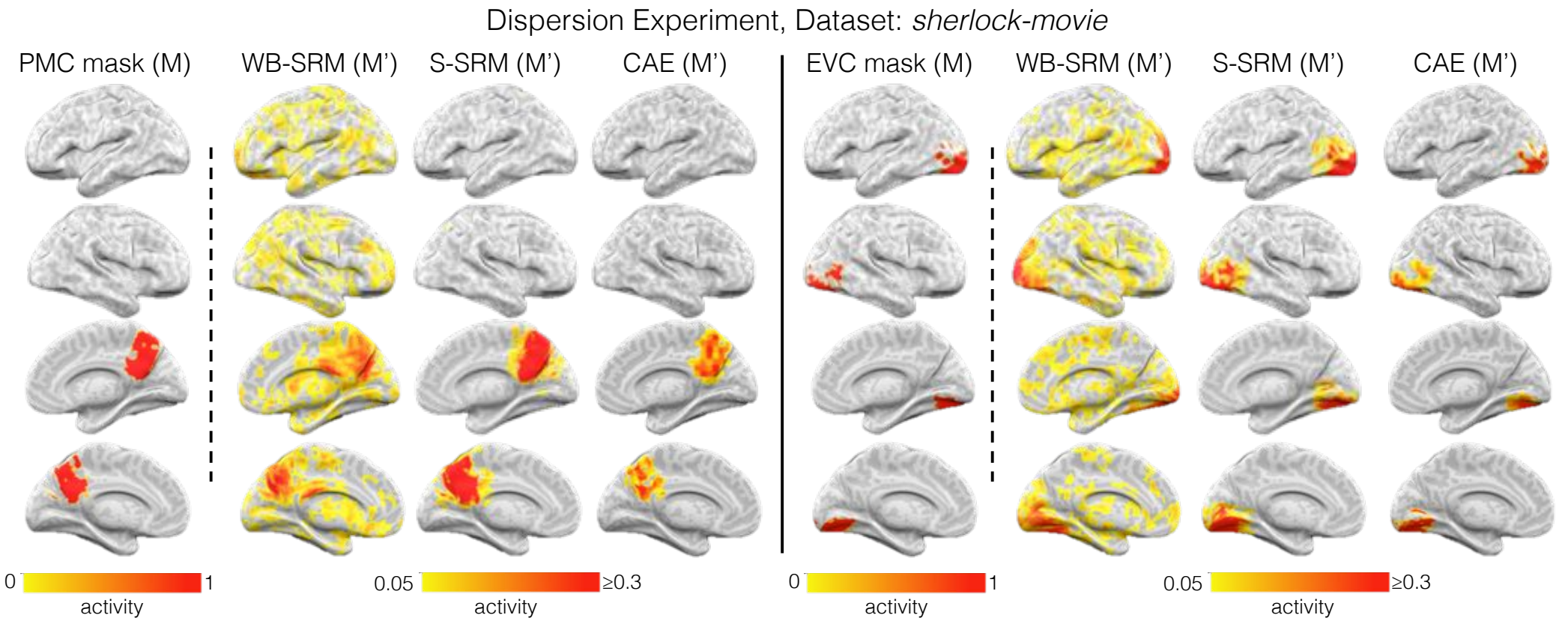

Figure 5: Experiment 4. A dispersion comparison between S-SRM and CAE using two anatomical ROI masks: Posterior Medial Cortex (PMC) and Early Visual Cortex (EVC).

The results of the time segment matching experiments indicate that both S-SRM and CAE enable improved matching of temporal segments over standard searchlight analysis (locally and globally) and marginally over WB-SRM (globally) (Fig. 3 (top left) and Fig. 4 (left)). For the *sherlock-movie* both S-SRM and CAE result in regions of highest accuracy in early visual cortex (EVC), which accords with neuroscience expectations for having early sensory areas driven in a specific and predictable way by the stimulus of the corresponding datasets. On the *sherlock-movie* dataset the peak predictive performance averaged across the top 0.5% ($\approx 350$ voxel locations) of the local regions is $0.11$ for SL, $0.35$ for S-SRM, and $0.14$ for the CAE. The averaged peak accuracy of the local regions of S-SRM and CAE clearly outperform those of standard SL analysis. S-SRM has done the best job of local aggregation of information areas with high peak accuracy in EVC. The CAE is second in rank with lower peak accuracy but nevertheless good coverage of relevant brain areas. Its peak accuracy is also in EVC. Our claim is also supported by the results of whole brain classification (Fig. 4) where S-SRM and CAE attain the highest classification accuracies; distinctly above standard searchlight analysis. For *audiobook*, both S-SRM and CAE have comparable spatial performance with highest accuracy in early audio cortex (EAC). The peak predictive performance averaged across the top 0.5% ($\approx 350$ voxel locations) of the local regions is $0.07$ for SL, $0.10$ for S-SRM, and $0.08$ for the CAE. While the averaged peak accuracy of local regions of S-SRM and CAE are slightly higher than SL analysis, the combined whole brain predictive accuracy (Fig. 4(left)) for both S-SRM and CAE are about twice as large as the best local region (Fig. 3 (bottom)).

It is particularly interesting that whole brain SRM (WB-SRM) does not perform as well as either of the local methods (S-SRM, CAE) when classifying temporal segments using whole brain data. This suggests that for cognitive state classification, it may be better to perform a local spatial analysis first, then combine the results of the local analyses to perform a global prediction of cognitive state.

The scene recall matching experiment provides a challenging task for all methods. We observe no improvement in whole brain classification over the best local prediction accuracies. This suggests that the relevant information is highly spatially localized. All three local analysis methods indicate that it is localized in the PMC ROI consistent with the finding in [36]. A follow-up experiment based only on PMC (Fig. 4 (right)) shows that the performance of the standard searchlight method and the CAE is the same when applied on the whole brain and when applied on PMC. On the other hand the performance of SRM and S-SRM improves when restricted to PMC. This suggests that with prior knowledge of informative local regions, it's best to use SRM and S-SRM directly in the ROI.

A key distinction between scene recall matching and time segment matching is that the scene recall test probes representations at a higher level of stimulus processing. It is known that neural representations become more abstract at higher and higher levels in the processing stream (e.g., as one moves from early sensory areas up to areas like PMC) [38]. Responses in higher level areas are generally less similar across subjects, compared to early sensory regions, likely due to their intrinsically more complex relationship to the stimulus; this property is observed in the identification of low-level sensory areas EVC (for *sherlock-movie*) and EAC (for *audiobook*) as some of the most informative voxels in the time segment matching test (Fig. 3). When two data types are not matched in terms of sensory input, these low-level areas do not match; instead, we find that PMC carries the most strongly shared information (the scene recall matching test). PMC is at the highest level of the stimulus processing stream, and as our experiments have displayed, it has the interesting property of exhibiting similar response patterns for two scenes with similar content irrespective to the type of sensory input (movie vs. spoken recall) [36]. Successful decoding of cognitive state using local information in the brain helps determine a local brain region's specific cognitive function, and also demonstrates what kind of information is present and the information's distribution across the brain. S-SRM and CAE have shown increased sensitivity for both local and global investigation.

In summary, we have investigated and compared two ways of preserving spatial locality in mutli-subject fMRI data aggregation: searchlight SRM and a convolutional autoencoder. Both approaches show improved results over standard competing methods. To our knowledge the application of a convolutional autoencoder to this task is novel and moves away from factor model approaches which appear to be hitting a performance ceiling. With further refinement, a well-trained convolution autoencoder may lead to a more powerful means of accomplishing the fMRI data aggregation task.

# A Convolutional Autoencoder for Multi-Subject fMRI Data Aggregation
## Supplementary Material

**Po-Hsuan Chen[1], Xia Zhu[2], Hejia Zhang[1], Javier S. Turek[2], Janice Chen[3], Theodore L. Willke[2], Uri Hasson[3], Peter J. Ramadge[1]**
[1]Department of Electrical Engineering, Princeton University
[2]Intel Labs
[3]Princeton Neuroscience Institute and Department of Psychology, Princeton University

### S.1 Notation

| Variable | Description |
|---|---|
| $v_x$ | number of voxels in $x$ dimension |
| $v_y$ | number of voxels in $y$ dimension |
| $v_z$ | number of voxels in $z$ dimension |
| $v$ | number of voxels , $v = v_x v_y v_z$ |
| $v_s$ | number of voxels in a searchlight |
| $i$ | index for subject, $i \in \{1, \dots, m\}$ |
| $t$ | index for TR, $t \in \{1, \dots, d\}$ |
| $q$ | index for feature, $q \in \{1, \dots, k\}$ |
| $X_i$ | observations from subject $i$<br>CAE: $X_i \in \mathbb{R}^{v_x \times v_y \times v_z \times d}$, WB-SRM: $X_i \in \mathbb{R}^{v \times d}$<br>SL-SRM: $X_i \in \mathbb{R}^{v_s \times d}$ for each searchlight |
| $S$ | estimated shared response<br>CAE: $S \in \mathbb{R}^{v_x \times v_y \times v_z \times k \times d}$, WB-SRM: $S \in \mathbb{R}^{k \times d}$<br>SL-SRM: $S \in \mathbb{R}^{k \times d}$, for each searchlight |

### S.2 SRM without orthogonality constraint is equivalent to a tied-weights linear fully-connected multi-view autoencoder with one hidden layer

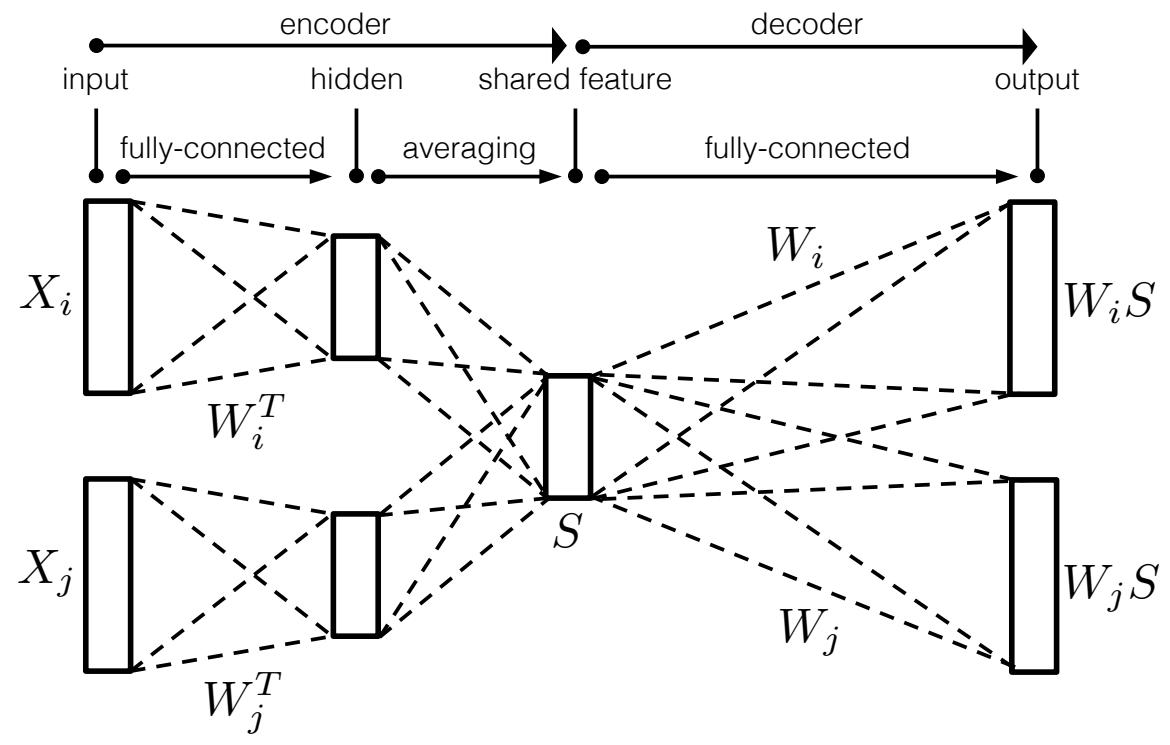

Figure 1: tied-weights linear fully-connected multi-view autoencoder

Since SRM does not keep track of spatial locality, the fMRI data is formulated by reshaping the 4D tensor response $(v_x, v_y, v_z, d)$ into 2D matrix $(v, d)$. fMRI time-series data $X_i \in \mathbb{R}^{v \times d}$, $i = 1{:}m$, is collected for $m$ subjects.

The core of SRM[1] can be viewed as

$$\begin{array}{ll} \min_{W_i,S} & \sum_i \|X_i - W_i S\|_F^2 \\ \text{s.t.} & W_i^T W_i = I_k, \end{array} \tag{1}$$

where $\|\cdot\|_F$ denotes the Frobenius norm. (1) can be solved iteratively by first initialize $W_i$, $i = 1{:}m$, and optimizing (1) with respect to $S$ by setting $S = {}^1\!/_m \sum_i W_i^T X_i$. With $S$ fixed, (1) becomes $m$ separate Procrustes problem [2] of the form $\min \|X_i - W_i S\|_F^2$ with solution $W_i = \tilde{U}_i \tilde{V}_i^T$, where $\tilde{U}_i \tilde{\Sigma}_i \tilde{V}_i^T$ is SVD of $X_i S^T$ [3]. These two steps iterate until a stopping criterion is satisfied.

SRM without the orthogonality constraint is equivalent to a tied-weight fully-connected linear multi-view autoencoder with one hidden layer as in Fig .1. Input data $X_i$ of subject $i$ is transformed through subject specific fully-connected transformation $W_i$ landing as $W_i^T X_i$ as hidden representation. The shared feature $S = {}^1\!/_m \sum_i W_i^T X_i$ is the average of each subjects' hidden representation similar to the formulation in SRM. From the shared feature, the decoder part of the network are tied-weight fully-connected transformations reconstructing the input from the shared feature $S$. The objective function of the network is loss between original data and reconstructed data $\min_{W_i} \sum_i \|X_i - W_i S\|_F^2$. This network performs similar to SRM without orthogonality but worse than SRM with orthogonality in experiments similar to [1]. This result is consistent with [1], which states that dropping the orthogonality constraint of SRM leading to decrease in performance.

### S.3 Generalization to nonlinear multi-view autoencoder

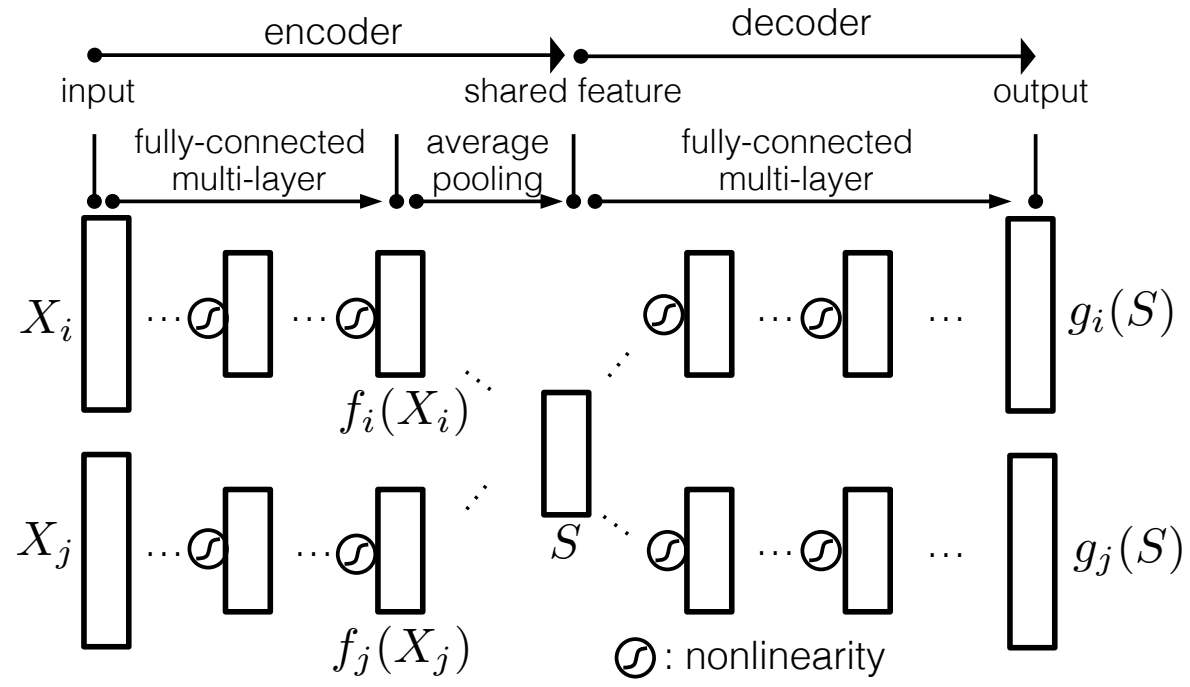

Figure 2: nonlinear fully-connected multi-view autoencoder

The linear fully-connected multi-view autoencoder, Fig. 1 can be generalized into a deeper network as in Fig. 2. We design a multi-layer multi-view auto-encoder. The encoders for each subject's data $X_i$ can be viewed as subject specific nonlinear function $f_i(\cdot)$, and subject. Each subject's response in feature space is $f_i(X_i)$, and shared feature $S = {}^1\!/_m \sum_i f_i(X_i)$ is the average across subjects. From the shared feature $S$, the decoder network reconstructed original input through nonlinear function $g_i(S)$. The whole network can be written as:

$$\min_{f_i, g_i} \quad \sum_i \|X_i - g_i(\tfrac{1}{m} \sum_j f_j(X_j))\|_F^2 + \lambda D_{KL}(\rho \| \hat{\rho}). \tag{2}$$

The first term is the mean squared error between the reconstructed output $g_i(\frac{1}{m} \sum_j f_j(X_j))$ and each subject's data; the second term is the Kullback-Leibler (KL) divergence to a binomial distribution with parameter $\rho$ [4]: $D_{KL}(\rho \| \hat{\rho}) = \rho \log(\frac{\rho}{\hat{\rho}}) + (1 - \rho) \log(\frac{1-\rho}{1-\hat{\rho}})$ with $\rho$ the desired sparsity and $\hat{\rho}$ the mean sparsity of the activations in the layer. This regularizes the network by sparsifying the shared feature maps $S$. Dropout is used to reduce overfitting [5]. We use the hyperbolic tangent $\tanh()$ activation function since it yields shared feature maps with positive and negative values as in competing methods.arsity, and the dropout probabilities. We select the parameters $\rho$ and $\lambda$ using cross-validation and fix dropout on hidden layers to the typical value of 0.5 [5] and deactivate it on the input layer. This network leads to comparable performance as SRM in predictive experiments as in [1].

### S.4 Comparison of Convolutional Autoencoder with various different parameter settings

In this subsection, we analyze the variability of the CAE performance when using different hyperparameter values. For this purpose, we repeat the time segment matching experiment on the *audiobook* dataset. We set the hyperparameters to default values of $\rho = 0.75$, $\lambda = 1$, $k_1 = 20$, and $k_3 = 20$, and then we visualize the performance of the model by varying each of these hyperparameters at a time. Fig. 3 shows the accuracy results when having a different number of filters $k_1$ in the first convolutional layer. Fig. 3 depicts the variability in accuracy when modifying the number of filters $k_3$ for the shared feature maps (layer 3). The expected mean sparsity hyperparameter $\rho$ and different values of the regularizer $\lambda$ are evaluated in Fig. 4, respectively. While the number of filters used to extract features from each subject volumes is crucial to obtaining good performance, the other parameters seem less sensible to value variations. We should note, however, that this fact depends on the application.

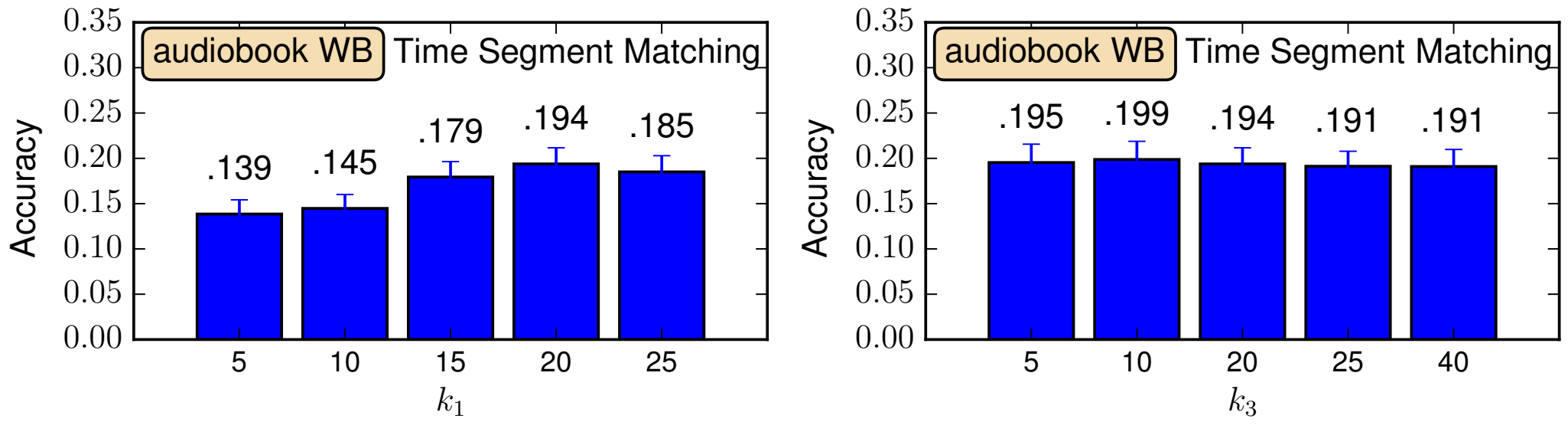

Figure 3: Accuracy variability of Convolutional Autoencoder on time segment matching experiment on the *audiobook* dataset for different (left) $k_1$ filters, (right) $k_3$ filters

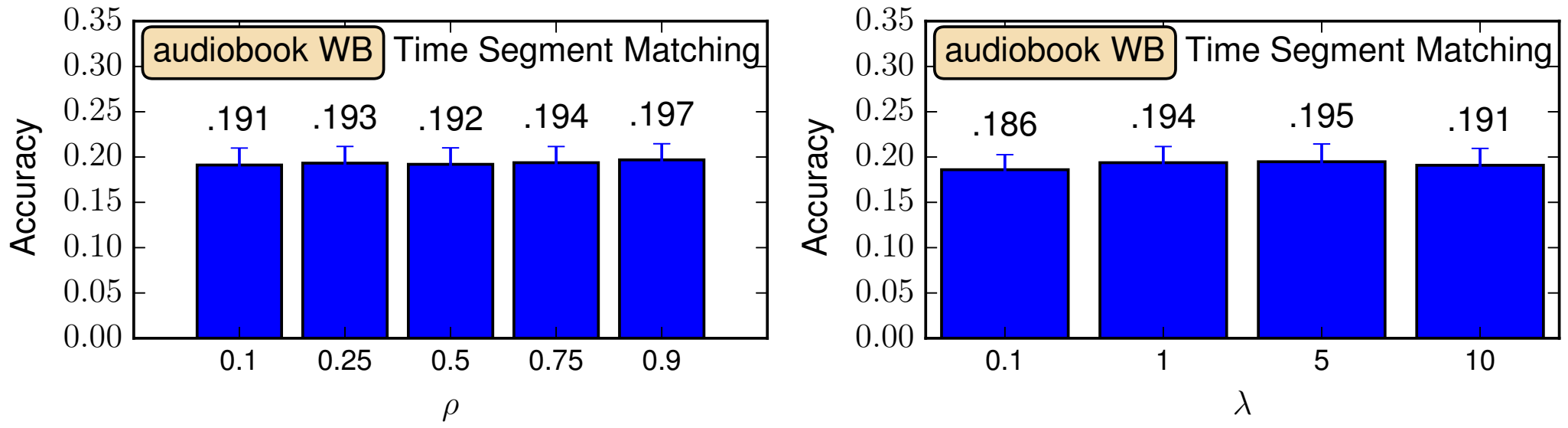

Figure 4: Accuracy variability of Convolutional Autoencoder on time segment matching experiment on the *audiobook* dataset for different (left) mean sparsity $\rho$, and (right) sparsity regularizer $\lambda$.

## S.5 Comparison on WB-SRM parameters

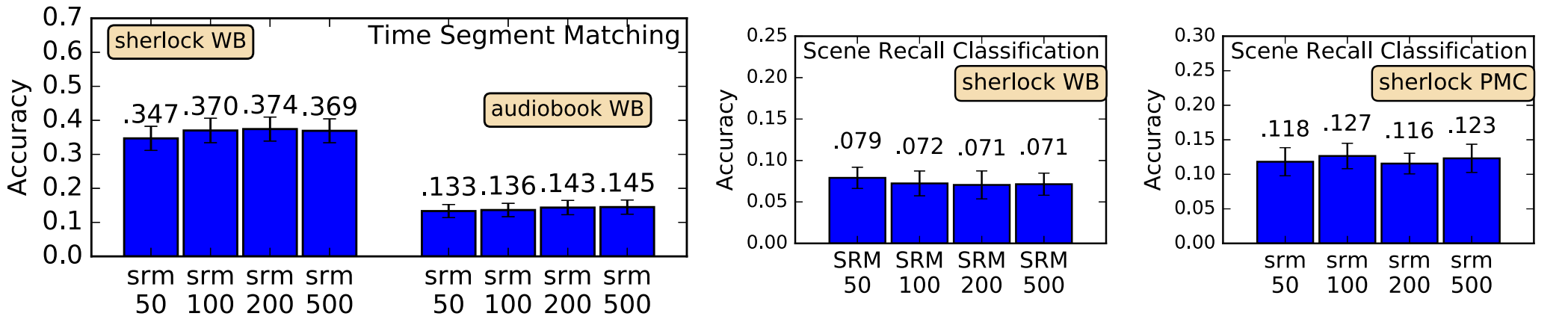

Figure 5: Comparison of SRM with different $k$. Left: Comparison of 9 TRs time segment classification on two datasets. Right: Comparison of movie scene recall classification on sherlock WB and PMC. Error bars: $\pm1$ stand. error.

## S.6 Comparison on S-SRM parameters

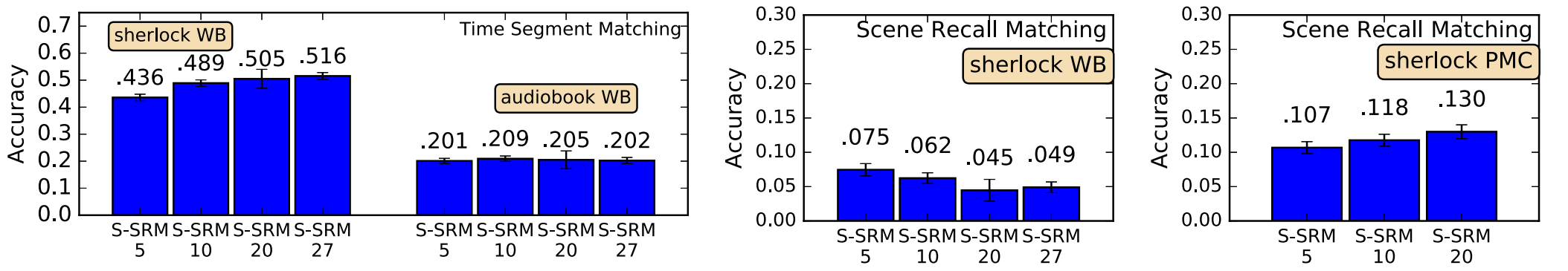

Figure 6: Comparison of S-SRM with different $k$. Left: Comparison of 9 TRs time segment classification on two datasets. Right: Comparison of movie scene recall classification on sherlock WB and PMC. Error bars: $\pm1$ stand. error.